\documentclass[runningheads, envcountsame, a4paper]{llncs}
\usepackage{threeparttable}
\usepackage{graphics}
\usepackage{times}
\usepackage{amssymb}
\usepackage{graphicx}
\usepackage[hyphens]{url}
\usepackage{array}
\usepackage{tabu}
\usepackage{tabstackengine}
\usepackage[nottoc]{tocbibind}
\usepackage[export]{adjustbox}
\usepackage{tikz}
\usepackage{multirow}

\usepackage{subfig}

\usepackage{graphics}
\usepackage{xcolor}

\def\XXX#1{\textcolor{red}{XXX #1}}
\def\fXXX#1{\footnote{\XXX{#1}}}
\def\XXX#1{}
\def\fXXX#1{}

\def\Sref#1{Section~\ref{#1}}
\def\Tref#1{Table~\ref{#1}}
\def\Fref#1{Figure~\ref{#1}}

\def\JRTk{JRTk}
\def\Janus{\JRTk{}}

\usepackage{amssymb}
\usepackage{pifont}
\newcommand{\yes}{\ding{51}}%
\newcommand{\no}{\ding{55}}%

\def\citet#1{\cite{#1}}

\newcommand{\keywords}[1]{\par\addvspace\baselineskip
\noindent\keywordname\enspace\ignorespaces#1}

\begin{document}


\title{A Speech Test Set of Practice Business Presentations\\ with Additional Relevant Texts}

\titlerunning{A Speech Test Set of Practice Business Presentations\\ with Additional Relevant Texts}

\def\orcidID#1{}
\author{Dominik~Mach{\' a}{\v c}ek\orcidID{0000-0002-5530-1615}  \and 
Jon{\' a}{\v s}~Kratochv{\' i}l\orcidID{0000-0002-9419-0241} 
\and 
Tereza~Vojt{\v e}chov{\' a}\orcidID{0000-0003-0245-8149} 
\and 
Ond{\v r}ej~Bojar\orcidID{0000-0002-0606-0050}    
}


\authorrunning{Dominik~Mach{\' a}{\v c}ek~et~al.}

\institute{ 
Charles University \\ 
Faculty of Mathematics and Physics \\
Institute of Formal and Applied Linguistics \\
Malostransk{\' e} n{\' a}m{\v e}st{\' i} 25,
118 00 Prague, Czech Republic \\
\email{$\{$machacek,jkratochvil,vojtechova,bojar$\}$@ufal.mff.cuni.cz} \\
}

\index{Mach{\' a}{\v c}ek, Dominik}
\index{Kratochv{\' i}l, Jon{\' a}{\v s}}
\index{Vojt{\v e}chov{\' a}, Tereza}
\index{Bojar, Ond{\v r}ej}

\toctitle{A Speech Test Set of Practice Business Presentations with Additional Relevant Texts} 
\tocauthor{Dominik~Mach{\' a}{\v c}ek, 
Jon{\' a}{\v s}~Kratochv{\' i}l,
Tereza~Vojt{\v e}chov{\' a},
Ond{\v r}ej~Bojar
}

\maketitle

%
%
%
%
\begin{abstract}
We present a test corpus of audio recordings and transcriptions of presentations of students' enterprises together with their slides and web-pages. The corpus is intended for evaluation of automatic speech recognition (ASR) systems, especially in conditions where the prior availability of in-domain vocabulary and named entities is benefitable. 
The corpus consists of 39 presentations in English, each up to  90 seconds long. 
The speakers are high school students from European countries with English as their second language.
We benchmark three baseline ASR systems on the corpus and show their imperfection.

\end{abstract}

\section{Introduction}

Nowadays, English is being widely used as lingua franca for communication between people without common first language (denoted as L1).
Europe is populated by dozens of nations with various and unique languages. In need for cooperation or interaction,
English is often \XXX{being}\XXX{JK:mozna bez toho being? Subjektivne mi to zni lip kdyz je tam jen used; Tea: me prijdou ok obe varianty; DM: mazu being} used as a universal first foreign language (or, in other words, the second language a human learns, L2) even between neighboring nations with closely related national languages, e.g. Czech and Polish. 
At the same time, many people are still not capable of using English and are
dependent on translation services, which in turn often rely on human experts. We
see an opportunity to boost availability, speed and language coverage of skilled
professional translators and interpreters with the help of machines. 

In spoken communication, such as during business conferences, the translation relies on speech comprehension. 
In Europe there are as many varieties of L2 English as there are European
languages because many speakers have an accent derived from their L1.
Current commonly used corpora for training the ASR systems are often based on audio recordings
of English L1 speakers \cite{tedlium,Librispeech}, which may not be optimal for
ASR of European L2 English. Furthermore, the outputs of ASR systems to date heavily
depend on domain coverage of training data and they could be considerably
improved by domain adaptation techniques. 
Also, the pronunciation of named entities from primarily non-English speaking areas usually differs significantly between English L1 and L2 speakers. Big corpora of L1 speakers often do not cover these differences and named entities are a big source of ASR errors and misunderstandings. \XXX{JK: misclassification mozna lepsi? Tea: zalezi, co chceme rict, jestli to, ze ASR chybuje a nerozumi anebo, ze zarazuje slova spatne.. nebo znamena misclassification jeste neco jineho? DM: ja chtel rict obecne "neporozumeni", to znamena jednak to, ze se neco prelozi spatne, tak taky ze vubec}

In certain situations, it is possible to prepare the ASR or spoken language translation (SLT) system for
the specifics of a given talks and speakers. This is due to the fact that the sessions such as conferences and meetings are often planned ahead of time and additional relevant materials such as accompanying presentations to the talks or relevant websites are available.

With this in mind, we have created a corpus consisting of practice presentations of student\XXX{DM: zadny "student's", to znamena "studentova", ani "students'", tj. "studentu", ale "student", coz je podstatne jmeno pouzite jako pridavne, takze studentsky. Navic ANTRE to tak taky pouziva.}  fictional firms. The corpus contains audio recordings, transcriptions and additional relevant texts (presentation slides and web pages) of the participants. The audio recordings cover English L2 speakers with eight European L1s (cs, sk, it, de, es, ro, hu, nl, fi). Some of the practise firms' web pages are in English, some of them in local languages. 
Our corpus is suitable for evaluation of ASR systems, both in settings with and without additional materials provided ahead of time.


In \Sref{sec:methodology}, we describe the methodology that was used to collect the corpus data.
In \Sref{sec:corpus}, we describe the corpus and its possible applications for the ASR systems. In \Sref{sec:asr-evaluation}, we present evaluation on three distinct 
English ASR systems.
We summarize related works in \Sref{sec:relevant} and conclude in \Sref{Conclusion}.

\section{Methodology}
\label{sec:methodology}

In this section, we explain the methodology we followed when creating the corpus. We collected the 
data at an international trade fair of student firms (see \Sref{sec:origin-of-data}), during a competition of business presentations (\Sref{sec:competition}). We motivated the speakers to transcribe their own speech presentations by introducing the Clearest voice competition for valuable prizes (\Sref{sec:clearest-voice}). Additionally, we collected documents related to the student firms (\Sref{sec:additional:resources}). Throughout the corpus creation, we adhered to ethical standards (\Sref{sec:ethical}).

\subsection{Background of Data: Student Firms and Trade Fair}
\label{sec:origin-of-data}

``Student firms'' are mock companies established for the practice of running a real company. The participants\XXX{,DM: tady nema bejt carka, viz {https://www.helpforenglish.cz/article/2006070601-vztazne-vety-relative-clauses}, je to vztazna veta urcujici} who run the companies are high-school
students, mainly from economically-oriented schools or departments. The firms meet at trade fairs, where they practise 
promoting their fictional goods or services, issuing invoices for mock trades, and bookkeeping. They also compete in aforementioned tasks and are evaluated based on various criteria by field professionals. The best firms advance into higher rounds of trade fairs, from regional rounds through national into international.


We collected the data at an international trade fair held recently in the Czech
Republic. The firms involved in our data collection were from 7 European
countries. See \Tref{tab:countries} for a summary.

\begin{table}[t!]
\centering
\caption{Number of student firms included in corpus and their countries of origin.
}
\label{tab:countries}
\small
\begin{tabular}{lc}
\textbf{Country} & \textbf{Firms} \\ 
\hline
Czech Republic &  18      \\
Italy & 8 \\
Romania & 4 \\
Slovakia & 3 \\
Austria & 2 \\
Spain & 2 \\
Belgium & 2\\
\hline
Total & 39 \\
\end{tabular}
\end{table}


The trade fair organizers provided us the firms' presentation slides, which were used by students during the fair. In many cases, we were able to find their web pages and included them into the corpus. See \Sref{sec:additional} for more details.



\subsection{Presentation Competition}
\label{sec:competition}



One of the activities during the fair, in which students could participate, was a competition of mock presentations of their businesses.
The subject of the competition was to promote the firm to a random stranger in
an elevator. The maximal allowed duration of the presentation was 90 seconds and
no additional materials were allowed to be shown. The participants had to use English and either one or two students were allowed to give the presentation.
A professional three-member committee was evaluating the content considering various aspects of the presentation. The selected competition winners were awarded prices for their performances.


We equipped the speakers with headset microphones to ensure the best possible
quality of recordings. Despite of that, there was loud background noise that
leaked to the recordings. On the one hand, this adds an extra obstacle for ASR, but on the other hand, 
the recordings thus represent a real environment where humans interact.
\subsection{Manual Transcriptions}
\label{sec:clearest-voice}

In order to obtain manual transcriptions of all the recordings, we asked the
participants to transcribe their speech, given only their own recording. To motivate the
students, we presented the task as an additional competition for valuable prizes. 
The objective of this competition was to find out who has the ``clearest voice'' for ASR. We processed the recordings with 
English ASR systems, evaluated them and awarded the students based on their respective ASR recognition scores. See \Sref{sec:asr-evaluation} for more details.

\def\expl#1{\textit{#1}}
 
The quality of the transcription was one of the major factors of the competition
(together with clarity of speech) because the students had no access to any ASR
outputs and had to assume that anything could be recognized correctly. We therefore believe that the students had a strong incentive to provide as accurate transcripts as possible.
Furthermore, we reviewed all the transcriptions and edited them to include the missing parts, normalize punctuation and correct the misspellings, but for authenticity, we preserved the original grammar and vocabulary, even when it was not considered as standard English (e.g. \expl{massageses} as a plural of \expl{massage}, or \expl{botel}, pronounced as \expl{bottle}, as a term for a \expl{hotel on a boat}).
\XXX{JK: tahle informace mozna vypada lepe ve footnote nez kdyz hned navazuje na to kde rikame, ze by mely byt vsechny prepisy od studentu v co nejlepsi kvalite? DM: spis ne. Ja bych to nechal. Jen to do camera-ready musime fakt udelat a pak tu vetu prepsat.}

\subsection{Additional Resources}
\label{sec:additional:resources}

As mentioned above, the participants of this trade fair competed in various disciplines, which included also the preparation of slides and web pages for the fictional companies and their products.

Thanks to our close collaboration with the main organizer of the event, we were able to obtain additional materials, where available.
While none of these additional materials were directly used in the presentation competition,
\XXX{JK: mozna je trochu matouci, ze nekde pouzivame jako nazev souteze "Clearest voice" a nekde "90 seconds" - chtelo by to nejak standartizovat; Tea: souhlas, myslim, ze jsme se to snazili cele anonymizovat a nemyslim si, ze jsme nekde nahore zminovali, ze to bylo 90 seconds soutez -> v nadpisu máme "Presentation Competition".. tak toho bych se asi drzela. DM: +1}
they were closely linked to the mock companies and their activity subject.
More details on the obtained and processed collection are available in \Sref{sec:additional}.

We are confident that the students did their best when preparing these materials, motivated by the various competitions.
For the purposes of ASR adaptation, the practical usability and overall quality of these materials highly differ from company to company. The relevant topics and named entities for each company are nevertheless mentioned in the corresponding materials.


\subsection{Ethical Standards}
\label{sec:ethical}


During the competition and subsequent data evaluation we did comply with the ethical standards, which are in Europe given by General
Data Protection Regulation (GDPR). Before the competition has started, all the
participants gave us their consent to use and release collected data for
research purposes, except of their names and any other personal data. Therefore,
we removed the real names of students from the recordings, transcriptions and
additional materials, and their photographs from the slides and downloaded web pages.


\section{Corpus}
\label{sec:corpus}
The main motivation for collecting the corpus was to test our current ASR models and to gather data for further improvement of their robustness. We believe
that the audio recordings contained in the corpus can be beneficial for anyone
who wants to deploy their ASR models in real world applications. We also believe that the model performance on these data is a good approximation of its general accuracy in noisy environment.

\subsection{Audio Recordings and Transcriptions}
\label{sec:audio}

The corpus consists of 39 recordings of presentations of fictional student firms. 
The content of the audio recording corpus is summarized in \Tref{tab:corpus-description} and the native languages of the speakers in \Tref{tab:corpus-native}. 

\begin{table} 
\begin{center}
\caption{Audio and content of the corpus.
}
\label{tab:corpus-description}
\small
\begin{tabular}{l@{~~}r@{~~}r@{~~}|@{~}r}
  {} & Single speaker  & Two speakers  & Total \\
\hline 
   Number of recording         &  17  &  22  & 39 \\
   Total audio duration         &  24m 20s  &  24m 8s & 58m 28s \\
   Transcription words         &  2891  &  3722 & 6613\\
   Distinct speakers         &   17  &  44  & 61 \\
\end{tabular}
\end{center}
\end{table}

\begin{table} 
\begin{center}
\caption{Native languages of the speakers in corpus.}
\label{tab:corpus-native}
\small
\begin{tabular}{l@{~~}c@{~~}c@{~~}c@{~~}c@{~~}c@{~~}c@{~~}c@{~~}c@{~~}c@{~~}|cr}
Language	& cs	& de	& it	& es	& ro	& sk	& hu	& nl	& fi & Total \\
\hline
Single speakers  	& 9	  & - 	& - 	& 1 	& 3 	&  1	& 1    	& -    	& -	& 17 \\
Two speakers     	& 18  &  4	&  16	&  2	&  -	& - 	& -    	& 3	& 1 &  44 \\
\end{tabular}
\end{center}
\end{table}

Recordings contain different types of background noise including live music,
announcements by organisers of the fair at main stage, conversations in different languages and noise produced by attendees of the presentations. 


%

\subsection{Topics}
\label{sec:topics}



The mock firms involved in the corpus represent a large variety of small or
medium-sized companies. We summarize their business fields in
\Tref{tab:business-branches}. The most common are travel agencies followed by
various food or beverage producers. Each firm is unique, focusing on a very
specific segment of the market. Most of the firms fictionally operate only in their local areas. 

\begin{table}[h]
\begin{center}
\caption{Business categories of student firms included in the corpus.}
\label{tab:business-branches}
\small
\begin{tabular}{lr}
 \textbf{Business Category} & \textbf{~Firms}  \\
\hline
   Travel agencies  & 7  \\
   Food and beverage producers & 4  \\
   Beauty and health & 3  \\
   Clothes and shoes & 3  \\
   Household equipment  & 3  \\
   Online promotion  & 2  \\
   Accessories & 2  \\
   Logistics & 2  \\
   Others & 13  \\
   \hline
   Total & 39 
\end{tabular}
\end{center}
\end{table}

\subsection{Additional Resources}
\label{sec:additional}

We collected additional resources of 36 student firms participating in corpus
creation. We are including either their presentation slides, web page or both.
The numbers and types of resources are described in
\Tref{tab:additional-resources}. In total, the additional resources contain
97\,000 of words, with a total vocabulary size of 15\,000.

\begin{table}[t]
\centering
\caption{Types of additional materials and number of firms providing them.}
\label{tab:additional-resources}
\small
\begin{tabular}{cc|rllllllllll}
\textbf{Slides} & \textbf{Web} &  \textbf{Firms} \\
\hline
\yes &  \yes  & 20  \\
\yes &  \no   & 12  \\
\no &   \yes  &  4  \\
\no &  \no & 3  \\
\end{tabular}
\end{table}


In order to protect the privacy of participants, we remove their real names and
photographs, however, we preserve all facts that are related to the companies
themselves. These include real or fictious email addresses, phone numbers, websites and locations.

The resources included in the corpus come in three distinct formats: the original
(either Microsoft Office presentation format, or original web content format
such as HTML or pictures), XLIFF format generated by MateCat Filters
tool,\footnote{\url{http://filters.matecat.com/}} which is an XML-based format
preserving the original structure of the document and may be useful for translation of the content, advanced information extraction tools, proper sentence segmentation or word-sense disambiguation. We also provide a plaintext format, which we created from XLIFF simply by extracting textual data from the documents. We included plaintexts because they may be convenient for the corpus users, and originals and XLIFF files, because they contain the complete information about the original structure of documents.

\subsection{Additional Resources by Languages}

The slides are either in Czech, Slovak or English. The web pages are mostly in
national languages.
Two of them are in multiple parallel language
variants and two are in English only. Despite this fact, we believe they can still be valuable
resource for ASR or SLT improvement with English as a source. We believe that the named
entities or specific in-domain vocabulary of the spoken presentation, which could otherwise be left unrecognized, may be
inferred from these documents even automatically.

We provide the language counts of presentations and web pages in
\Tref{tab:additional-languages}.
We note that there is one company in the corpus whose
presenter's L1 was Hungarian, their slides were English and web page in Romanian.

All the documents in the corpus are marked with language tags.

%

\begin{table}[t]
\begin{center}
\caption{Languages of presentation materials}
\label{tab:additional-languages}
\small
\begin{tabular}{l@{~}c@{~}c@{~}c@{~}c@{~}c@{~}c@{~}c@{~}c@{~}c@{~}c@{~}c@{~}|r}
Lang.	& cs	& en	& de	& it	& es	& ro	& sk	& cs/en	& ro/en	& sk/en	& it/en/es/de	& total \\
\hline
Slides  	& 14	& 15	& - 	& - 	& - 	& - 	&  1	&     1	& -    	&     1	& -          	&    32 \\
Web     	& 14	&  2	&  2	&  2	&  1	&  1	& - 	& -    	&     1	& -    	&           1	&    23
\end{tabular}
\end{center}
\end{table}



\section{Evaluation of ASR}
\label{sec:asr-evaluation}

In order to document the state of the art of ASR, we evaluated three ASR systems on the corpus. 




\subsection{The ASR Systems}
\label{sec:asr-systems}

We consider three different ASR systems:

\textbf{Janus Recognition Toolkit (\Janus{})}
\cite{Janus-toolkit} featuring the IBIS single pass decoder \cite{ibis}. Its acoustic model was trained on TED talks \cite{tedlium} and Broadcast News \cite{broadcast-news}. This system was designed to recognize lecture talks from IWSLT 2017 workshop \cite{janus-iwslt}.

\textbf{Google Cloud Speech-to-Text\footnote{\url{https://cloud.google.com/speech-to-text/}}
}
with English (United States) language option.\XXX{DM: Je to tak, Jonasi?}

\textbf{Kaldi} \cite{Kaldi} based model
trained on data from Multi-Genre Broadcast Challenge \cite{mgb_challenge}, on 1600 hours of broadcast audio from BBC TV and several hundred million words of subtitle text for language modeling. This model is thus suitable mainly for native British English speakers.

We also tried Microsoft Cloud ASR but it failed for all our recordings.

\begin{table}[t]
\begin{center}
\caption{WER of JRTk, Kaldi BBC and Google model scores on all recordings in the
corpus (right) and on the recordings on which all the systems produced some
output (left). WER of 100\% indicates that no output was provided.
}
\label{tab:all-data-scores}
\small
\begin{tabular}{l@{~~}|r@{~~}r@{~~}r@{~~}|r@{~~}r@{~~}r@{~~}rrr}
{} & \multicolumn{3}{c|}{\textbf{Recognized by all}} & \multicolumn{3}{c}{\textbf{All recordings}}\\
  {} & Google  & Kaldi B. & JRTk & Google & Kaldi B. & JRTk \\
\hline 
   Mean         & 73.59 & 87.55 & 45.21&  89.32  &  87.47  &  45.63          \\
   Min         & 20.90 & 83.96 & 25.00 &  20.90  &  83.96  &  25.00           \\
   Max         & 98.31 & 91.03 & 74.08 &  100.00  &  91.03  &  99.58          \\
   Median       & 87.50 & 87.59 & 43.41  &  100.00  &  87.04  &  46.31       \\
   Stddev    & 27.87 & 2.29 & 15.28     &   21.82  &  1.92  & 15.23  \\
\end{tabular}
\end{center}
\end{table}

\subsection{Evaluation Metric}
\label{sec:eval-wer}
We use the standard word error rate (WER) metric, which is the minimum number of
text insertions, deletions and substitutions needed to transfer one document to
another, normalized by the total number of words in the document.
As customary in ASR development, we disregard letter case and punctuation for
this evaluation.
We took the transcriptions obtained from the participants as the ground truth against which the automatic speech recognition outputs were evaluated.

\subsection{Results}
\label{sec:eval-results}

\begin{figure*}
	    \centering
		\input{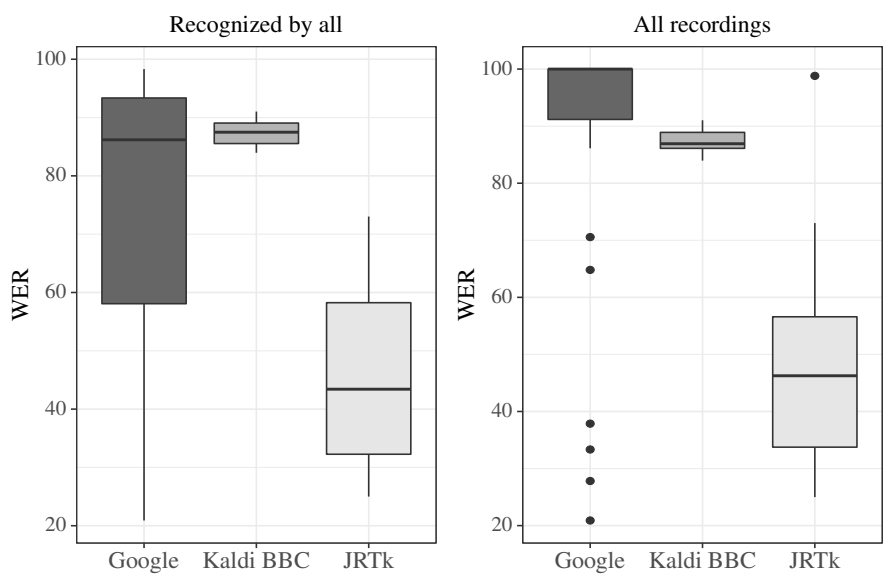}

		\caption{Boxplot showing the word error rate scores of Google,
		Kaldi BBC and JRTk models on all recordings (right) and on a subset
		where all the systems produced some non-empty output (left).
		}
		\label{plot:WER}
\end{figure*}

The descriptive statistics of respective word
error rate scores are listed in \Tref{tab:all-data-scores} and visualized in \Fref{plot:WER}. Note that the
lower WER, the better recognition.

As already discussed in \Sref{sec:corpus}, the audio files contain
a significant amount of background noise. Due to this fact, Google \XXX{podle toho, jakej to je} returned an empty output in some cases, resulting in the WER of
100\%. In order to account this, we selected only the recordings on which all the systems had less than 100\% WER, and measured a second set of
descriptive statistics on this subset. 


By manually inspecting the recordings on which the systems had the highest
error rate we observed that the ASR difficulties could have been caused by a
very strong accent of the speaker, or by the fact that the microphone was not in
the appropriate distance from the mouth, or that the speaker did not articulate
clearly. 
Also, the background conditions such as a music band playing or students entering the presentation room may have affected the recognition quality. 







\section{Related Works}
\label{sec:relevant}

\XXX{Dominik dela resersi.}

%
%


\XXX{OB: Dobry clanek se na prvni pohled pozna podle toho, ze cituje jak stare
zdroje, tak nejnovejsi zdroje. ;-) Muzeme tedy v resersi trochu blafovat. Jako
zacatek bych citoval https://arxiv.org/pdf/1808.05312.pdf (jak uz jsem psal
nahore) a z nej jeste nejake zajimave dalsi odkazy. V resersi bych se soustredil
na nejcerstvejsi clanky o speaker adaptation, L1 adaptation a domain adaptation.
Ale chapu, ze za pul nedelniho odpoledne se toho moc nestihnei, tak se tim moc
netrap.}

\XXX{DM: za mne asi hotovo.}

\def\subt#1{\textbf{#1}}


Tests sets for ASR are usually released together with speech corpora \cite{tedlium,Librispeech,euronews}. Our corpus is unique in a way that it contains L2 English, similarly as \citet{arctic}, but in our corpus there is a large variety of speakers, European L1s and realistic background noise conditions.
Also, to our best knowledge, there is not any other speech corpus with additional in-domain resources.

\subt{Robustness to noise:}
There are some corpora intended for noisy speech recognition: \citet{subjective,noisy-timit,chime,spine}.
In \citet{162k}, the authors show that model trained on a large data-set of distorted data with background noise is able to generalize much better than domain-specific models. Similar conclusions were derived in \citet{Artifitial-noise}, where the authors experimented with random sampling of noise and intentionally corrupting the training data. 

\subt{Non-native speech:}
Adaptation for non-native speech in low-resource scenarios was studied by \citet{acoustic-non-native}, who proposed interpolation of acoustic models or polyphone decision tree specialization. This can be incorporated into statistical ASR systems. 
For hybrid HMM-DNN (Hidden Markov models and deep neural networks) models, data
selection methods can be used. In \citet{svetlana}, combination of L2 out-of-domain read speech and L2 in-domain spontaneous speech led to the highest improvements, as opposed to using L1 speech.

\subt{Domain adaptation:}
For purely neural LF-MMI (Lattice-free maximum mutual information) models
\cite{lf-mmi}, multi-task learning with large out-of-domain data as a first task and in-domain data as a second task, or various approaches of transfer learning can be beneficial \cite{transfer}.

\section{Conclusion}
\label{Conclusion}
We presented a small English speech corpus (only about 1 hour in total) intended as a
test set for challenging speech recognition conditions: 61 distinct speakers,
none of which were native speakers of English, a diverse set of vocabulary domains
and noisy background.

We have demonstrated that current ASR systems have severe difficulties in processing
the test set, with WER ranging from 40 to 100\% on individual audio recordings.
The test set is equipped with additional text materials which can serve as evaluation of domain adaptation.

The corpus is publicly released and available under the following link: 
\begin{center}
    \url{http://hdl.handle.net/11234/1-3023}.
\end{center}

\section*{Acknowledgements}

This research was supported in parts by the grants H2020-ICT-2018-2-825460
(ELITR) of the European Union and 19-26934X (NEUREM3) of Czech Science
Foundation.

 We are grateful to the organization team of the fictional student firms fair, who allowed us to conduct the competition during the event. We are also grateful to the students, who presented their firm and transcribed their audio recordings. Last but not least we are thankful to the team in Karlsruhe Institute of Technology and to the PerVoice team, who helped us overcome the technical difficulties that we have encountered.

\bibliographystyle{splncs03}
\bibliography{paper}

\end{document}